\documentclass[acmtog]{acmart}
\AtBeginDocument{%
  }

\copyrightyear{2026}
\acmYear{2026}
\setcopyright{cc}
\setcctype{by}
\acmConference[SA Conference Papers '26]{SIGGRAPH Asia 2026 Conference Papers}{December 01--04, 2026}{Kuala Lumpur, Malaysia}
\acmBooktitle{SIGGRAPH Asia 2026 Conference Papers (SA Conference Papers '26), December 01--04, 2026, Kuala Lumpur, Malaysia}
\acmDOI{10.1145/3829340.3842249}
\acmISBN{979-8-4007-2842-6/2026/12}

\usepackage{booktabs}
\usepackage{multirow}
\usepackage{graphicx}
\usepackage{enumitem}
\makeatletter
\newcount\PIEPSAuthorLineCount
\let\PIEPSOriginalTypesetAuthorLine\@typeset@author@line
\def\@typeset@author@line{%
  \advance\PIEPSAuthorLineCount by 1\relax
  \ifnum\PIEPSAuthorLineCount=1
    \let\PIEPSOriginalAndWord\@@and
    \def\@@and{\unskip;}%
    \PIEPSOriginalTypesetAuthorLine
    \let\@@and\PIEPSOriginalAndWord
  \else
    \PIEPSOriginalTypesetAuthorLine
  \fi}
\makeatother
\begin{document}


\title{PIE-PS: Photometric Stereo from Physical Irradiance Event Streams}

\author{Xiangze Meng}
\affiliation{%
  \institution{Jilin University}
  \city{Changchun}
  \country{China}}
\affiliation{%
  \institution{Guangdong Institute of Intelligence Science and Technology}
  \city{Zhuhai}
  \country{China}}
\email{mengxiangze520@gmail.com}

\author{Guangyu Li}
\affiliation{%
  \institution{Guangdong Institute of Intelligence Science and Technology}
  \city{Zhuhai}
  \country{China}}
\email{liguangyu@gdiist.cn}

\author{Jing Li}
\affiliation{%
  \institution{Jilin University}
  \city{Changchun}
  \country{China}}
\email{lij25@mails.jlu.edu.cn}

\author{Di Mei}
\affiliation{%
  \institution{Beijing Institute of Technology}
  \city{Beijing}
  \country{China}}
\email{meidi@connect.hku.hk}

\author{Songchen Ma}
\affiliation{%
  \institution{The Hong Kong University of Science and Technology}
  \city{Hong Kong}
  \country{China}}
\email{songchenma@ust.hk}

\author{Mingkun Xu}
\authornote{Corresponding authors.}
\affiliation{%
  \institution{Guangdong Institute of Intelligence Science and Technology}
  \city{Zhuhai}
  \country{China}}
\email{xumingkun@gdiist.cn}

\author{Rui Ma}
\authornotemark[1]
\affiliation{%
  \institution{Jilin University}
  \city{Changchun}
  \country{China}}
\email{ruim@jlu.edu.cn}

\renewcommand{\shortauthors}{Meng et al.}

\begin{abstract}
Event cameras record asynchronous log-image-irradiance changes with microsecond latency and high dynamic range.
These properties are useful for photometric stereo under moving illumination, but raw events are sparse and depend on an unknown contrast threshold.
We start from the event trigger model and derive a physical relation between adjacent events, light motion, and surface normals.
This relation gives a direct physics-only solver, but the solver needs the threshold, enough events at each pixel, and independent per-pixel optimization.
To address these limits, we introduce PIE-PS, a learning-based framework for dense surface normal reconstruction from raw event streams and known lighting.
We form Physical Irradiance Events (PIEs) by pairing two adjacent events at the same pixel with their corresponding light directions.
Each PIE provides a Physical Irradiance Event Feature (PIEF), defined as the signed event rate.
PIEF does not require the unknown contrast threshold.
To share spatial and temporal context across nearby PIEs, we introduce PIE-GNN, which treats each PIE as a graph node and encodes it with its light-pair geometry. Since the reliability of PIE observations can vary with local appearance, illumination geometry, and sensor noise, Reliability-Grading Attention (RGA) predicts reliability weights to down-weight unreliable PIEs. Pixel aggregation then produces dense normals.
Experiments on synthetic and real data show that PIE-PS outperforms prior event-based photometric stereo methods and the direct solver baseline.
Project page: \url{https://pie-ps.github.io}.
\end{abstract}

\ccsdesc[500]{Computing methodologies~Reconstruction}

\keywords{Photometric Stereo, Event Camera, Physical Irradiance Events}
\begin{teaserfigure}
  \includegraphics[width=\textwidth]{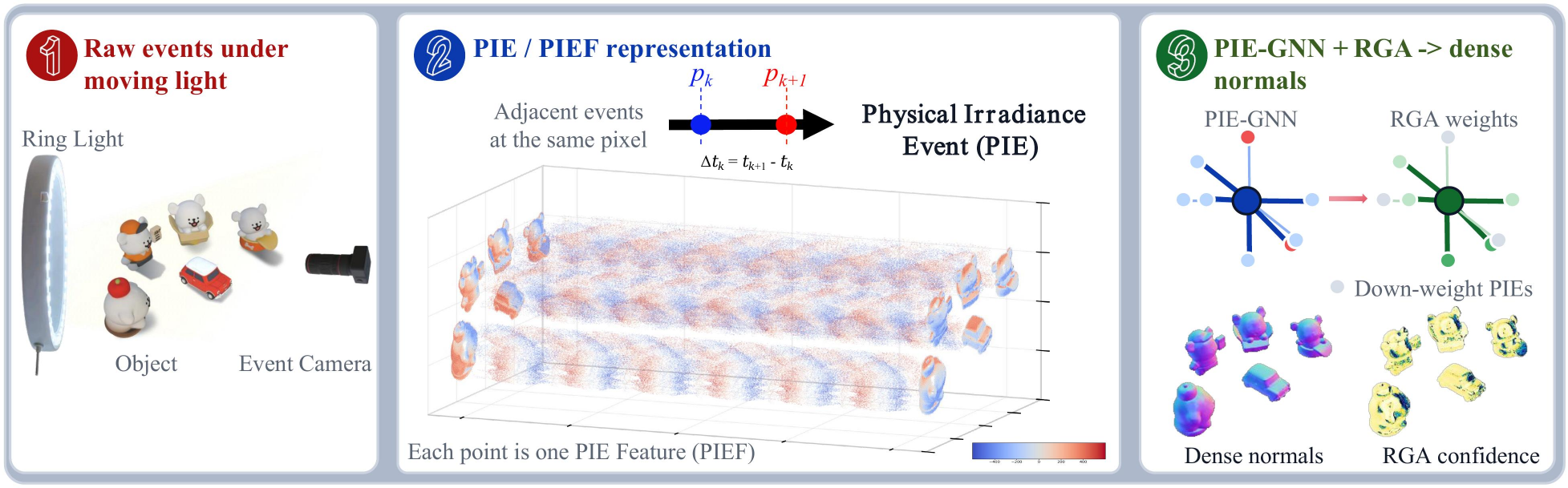}
  \caption{
  Overview of PIE-PS.
  A prototype system captures raw event streams under moving light.
  Adjacent events at the same pixel are paired with their light directions to form Physical Irradiance Event (PIE) observations.
  Each PIE provides a threshold-free signed event-rate Physical Irradiance Event Feature (PIEF) and calibrated light-pair geometry.
  PIE-GNN encodes these PIE nodes, Reliability-Grading Attention (RGA) down-weights unreliable observations, and pixel aggregation produces dense surface normals.
  }
  \Description{Overview of PIE-PS from raw event capture under moving light to PIE representation, graph-based RGA weighting, and dense normal reconstruction.}
  \label{fig:teaser}
\end{teaserfigure}


\maketitle

\section{Introduction}

Recovering detailed surface geometry is important for content creation, inspection, and augmented reality~\cite{qi2018geonet, bae2024rethinking}.
Photometric Stereo (PS)~\cite{woodham1980photometric} estimates dense surface normals from shading changes under varying light.
It can recover fine detail, but standard frame-based PS records one image for each light state.
This limits frame rate and dynamic range, especially under fast lighting or large intensity variation.

Event cameras provide a different sensing model.
They record asynchronous log-image-irradiance changes with microsecond latency and high dynamic range~\cite{event-based-vision, rebecq2019events}.
Under moving light, these events contain normal cues.
These properties make event-based PS more than a direct reuse of frame-based PS.

Prior analytical methods show that event-based PS is possible.
EventPS~\cite{eventps} formulates event-based photometric stereo as per-pixel reconstruction from changes in log image irradiance.
This connects event measurements with the lighting trajectory, but it depends on a fixed event contrast threshold.
PS-EIP~\cite{PS-EIP} improves robustness by using an interval-profile approximation over three recorded events.
However, this design requires a look-ahead event and does not use event polarity.
Their physical formulations are still built on local per-pixel event constraints.
Thus, the reconstruction can have limited spatial context and can be sensitive to event density, threshold calibration, and locally unreliable observations.

We start from the same event trigger model under moving light, but use it in a different way.
Prior analytical methods mainly turn the event stream into a per-pixel differential or interval-profile equation.
We instead derive a pairwise photometric constraint from two adjacent events and the light directions at their timestamps.
This gives a direct relation between the event pair, the light-pair change, and the surface normal.
If the contrast threshold is known and enough event pairs are available at a pixel, these constraints can be solved by direct numerical optimization.
We keep this solver as a physics-only baseline because it makes the observation model explicit.
It also shows the limits of a pure analytical route. The solver needs the unknown threshold \(C\), requires enough events to build stable equations, and treats each pixel independently.

Fig.~\ref{fig:teaser} gives an overview of the PIE-PS representation and reconstruction pipeline.
To address these limits, we introduce a learning-based framework built on Physical Irradiance Events (PIEs).
A PIE is not a new sensor event.
It is a structured observation built from two adjacent events at the same pixel and their corresponding light directions.
Its Physical Irradiance Event Feature (PIEF) is a scalar signed event rate, given by event polarity divided by inter-event time.
The light-pair geometry remains part of the PIE observation and is added when building the graph node.
We then treat each PIE as a graph node and use PIE-GNN to exchange information between nearby event-pair observations in space and time.
This graph form shares context across nearby pixels and times.
In practice, the reliability of PIE observations can vary with local appearance, illumination geometry, and sensor noise, while less reliable observations may still retain useful cues for normal estimation.
Reliability-Grading Attention (RGA) therefore assigns each PIE a soft reliability grade before pixel aggregation.
The weighted PIE embeddings are then aggregated into pixel features, and a lightweight pixel aggregator predicts dense surface normals.

Our method separates the roles of physics and learning.
The physical model tells us what an event pair should measure and which light-pair cues are useful.
The graph network handles the parts that are hard to solve analytically, including unknown threshold, uneven event density, varying PIE reliability, and spatial context.
As a result, PIE-PS uses the event formation model without making the fixed-threshold analytical solver the main reconstruction method.

Our main contributions are summarized as follows:
\begin{itemize}[leftmargin=*, itemsep=0pt, topsep=2pt]
    \item We formulate pairwise event photometry from adjacent event triggers and use it to build a direct numerical solver baseline.
    \item We introduce PIE-PS, a framework for event-based photometric stereo from raw event streams and known lighting.
    \item We develop PIEF, PIE-GNN, and RGA as the main learned method for threshold-free signed-rate encoding, light-pair contextual feature extraction, and unreliable-event suppression.
\end{itemize}

\section{Related Work}
\subsection{Photometric Stereo}
Photometric Stereo (PS) was pioneered by Woodham~\cite{woodham1980photometric} to recover surface normals from varying illumination.
To handle shadows, specularities, and other non-Lambertian effects, later methods use robust optimization, sparse regression, low-rank modeling, and benchmarked evaluation on DiLiGenT~\cite{DiLiGenT),op1,op2,non-lam1,non-lam2}.
Learning-based methods further map calibrated or uncalibrated image observations to normals~\cite{DPSN,SDPS,deep1,deep2,deep3,deep4,li2024spin,ge2025prm,lichy2022fast}.
Related reflectance-acquisition methods recover shape or normals together with polarimetric SVBRDFs~\cite{baek2018polarimetricSVBRDF,hwang2022sparseEllipsometry}.
These works target richer material capture, while our work focuses on calibrated event-based dense normal recovery under moving illumination.
Frame-based PS methods remain limited by sensor frame rate, motion blur, and dynamic range in high-speed lighting scenarios~\cite{deep5}.

\subsection{Event-based 3D Vision}
Event cameras offer high dynamic range and microsecond temporal resolution by asynchronously recording intensity changes.
Recent surveys summarize event-based 3D reconstruction under high-speed motion, low light, and high dynamic range scenes~\cite{xu2025event3dSurvey}.
Early event-based 3D methods focus on sparse depth estimation, multi-view stereo, structured light, and visual odometry~\cite{event-based-vision,emvs,Esl,niu2025esvo2}.
Other approaches use events for visual hull reconstruction~\cite{wang2022evac3d} or combine them with implicit representations, such as EventNeRF~\cite{rudnev2023eventnerf,feng2025ae}, event-based NeRF variants~\cite{nerf1,nerf2}, and Event-3DGS~\cite{han2024eventgs,feng20254dgs}.
These methods mainly target depth, trajectory tracking, novel view synthesis, or 3D reconstruction, rather than controlled-light pixel-wise normal estimation.

\subsection{Event-based Photometric Stereo}
Recent research has used event streams for high-speed normal estimation.
Hybrid methods, such as EFPS-Net~\cite{EFPS-Net}, fuse RGB frames with event data to enhance boundary and specular details.
EventPS~\cite{eventps} is a calibrated pure event-based method that formulates a per-pixel differential constraint on log image irradiance from the light trajectory, but it depends on the event contrast threshold and local event density.
PS-EIP~\cite{PS-EIP} improves robustness with an Event Interval Profile built from three events, although it requires a look-ahead event and does not use event polarity.
EventUPS~\cite{eventups} extends event-based PS to the uncalibrated illumination setting.
In contrast, we use adjacent-event pairwise photometry to build PIEs that preserve event timing, polarity, and light-pair geometry without requiring the contrast threshold at test time.
PIE-GNN then shares context across nearby PIE nodes, and RGA weights unreliable PIEs before pixel aggregation produces dense normals.

\section{Methodology}
\begin{figure*}[h]
  \centering
  \includegraphics[width= 0.95\textwidth]{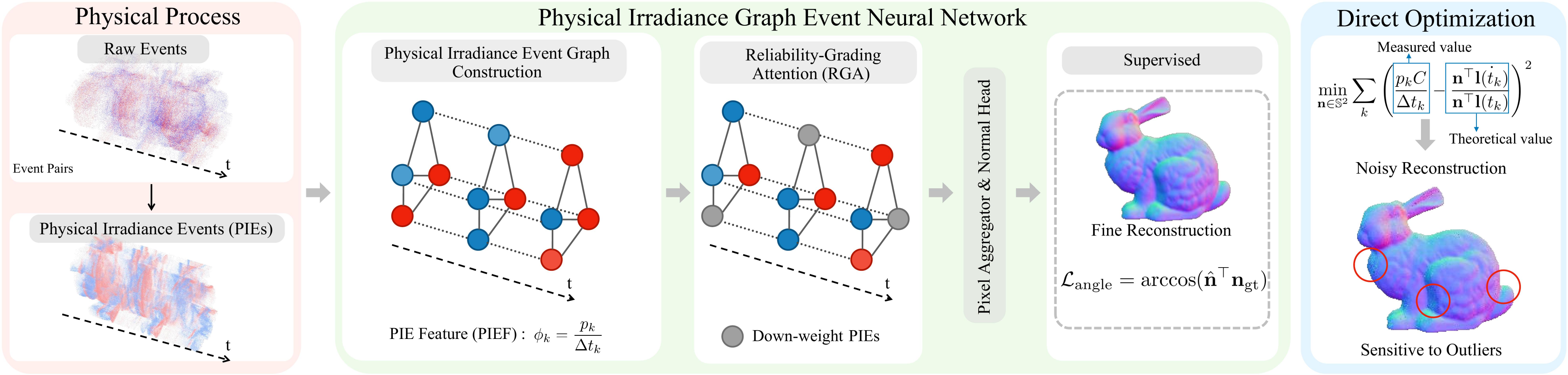}
  \caption{ Overview of the PIE-PS framework. We first convert raw events under moving light into PIE observations. Each PIE provides a threshold-free signed event-rate feature and a corresponding light-pair geometry. The PIE-GNN encodes these observations, uses Reliability-Grading Attention (RGA) to weight them, aggregates pixel features, and predicts normals with a pixel aggregator.}
  \label{pipeline}
\end{figure*}

We propose PIE-PS for surface normal recovery from event streams. Fig.~\ref{pipeline} shows the pipeline.
We first derive event-pair photometry and build Physical Irradiance Events (PIEs) from adjacent events and their light directions in Sec.~\ref{Theoretical Foundation}.
Each PIE provides a Physical Irradiance Event Feature (PIEF), defined as the threshold-free signed event rate of the event pair in Sec.~\ref{sec:network}.
PIE-GNN then uses the PIEF together with the corresponding light-pair geometry to extract node features in Sec.~\ref{sec:network}.
RGA weights unreliable PIEs, and pixel aggregation predicts dense normals as described in Sec.~\ref{sec:LCA}.

\subsection{Physical Modeling of Event Photometry}
\label{Theoretical Foundation}
We consider a pixel location $\mathbf u$ on a static surface.
Its unit surface normal is $\mathbf n_{\mathbf u} \in \mathbb S^2$.
The diffuse albedo is $\rho \in \mathbb R^+$.
Under a distant directional light $\mathbf l(t) \in \mathbb S^2$ with intensity $s$, the image irradiance at pixel $\mathbf u$ is
\begin{equation}
I(t)=s\rho\,\mathbf n_{\mathbf u}^{\top}\mathbf l(t),
\label{EQ:Lambertian}
\end{equation}
for lit regions where $\mathbf n_{\mathbf u}^{\top}\mathbf l(t)>0$.

An event camera measures changes in log image irradiance. We write the log image irradiance as
\begin{equation}
\mathcal L(t)=\ln(qI(t)),
\label{EQ:Log_irrandiance}
\end{equation}
where $q$ is a sensor gain.
Using Eq.~\ref{EQ:Lambertian}, we have
\begin{equation}
\mathcal L(t)
=
\ln q+\ln s+\ln\rho
+
\ln\left(\mathbf n_{\mathbf u}^{\top}\mathbf l(t)\right).
\label{EQ:log_lambertian}
\end{equation}
The first three terms are constant for a fixed pixel.
Taking the time derivative gives the theoretical value
\begin{equation}
\dot{\mathcal L}(t)
\triangleq
\frac{\partial \mathcal L(t)}{\partial t}
=
\frac{
\mathbf n_{\mathbf u}^{\top}\dot{\mathbf l}(t)
}{
\mathbf n_{\mathbf u}^{\top}\mathbf l(t)
}.
\label{GT}
\end{equation}
This value is defined by the normal and the light motion.

The event camera does not measure Eq.~\ref{GT} directly.
It reports an event when the accumulated log-image-irradiance change reaches a threshold.
At pixel $\mathbf u$, let two adjacent events be $e_k$ and $e_{k+1}$.
Their timestamps are $t_k$ and $t_{k+1}$, and the polarity of the second event is $p_{k+1}\in\{+1,-1\}$.
The trigger model gives
\begin{equation}
\mathcal L(t_{k+1})-\mathcal L(t_k)\approx p_{k+1}C,
\label{EQ:event_pair_trigger}
\end{equation}
where $C$ is the contrast threshold.
With $\Delta t_k=t_{k+1}-t_k$, the event gives a finite-difference measurement:
\begin{equation}
\dot{\mathcal L}(t_{k+1})
\approx
\frac{\mathcal L(t_{k+1})-\mathcal L(t_k)}{\Delta t_k}
\approx
\frac{p_{k+1}C}{\Delta t_k}.
\label{calculate}
\end{equation}

Eq.~\ref{GT} gives the theoretical value, while Eq.~\ref{calculate} gives the event-camera measurement.
Matching them gives the following pixel-wise objective:
\begin{equation}
\min_{\mathbf n_{\mathbf u}\in\mathbb S^2}
\sum_k
\left(
\frac{p_{k+1}C}{\Delta t_k}
-
\frac{
\mathbf n_{\mathbf u}^{\top}\dot{\mathbf l}(t_{k+1})
}{
\mathbf n_{\mathbf u}^{\top}\mathbf l(t_{k+1})
}
\right)^2 .
\label{EQ:direct_matching_objective}
\end{equation}
This objective can be solved directly by numerical optimization to recover the normal $\mathbf n_{\mathbf u}$.
However, using this objective as a reconstruction method leads to practical limitations.
It requires a fixed event contrast threshold $C$.
It also needs enough event-pair observations at each pixel to make the normal well constrained.
These limitations motivate our Physical Irradiance Event (PIE) representation.
A PIE is not a new sensor event.
It is a paired observation built from two adjacent events at the same pixel.
Let the raw events be $e_k=(\mathbf u,t_k,p_k)$ and $e_{k+1}=(\mathbf u,t_{k+1},p_{k+1})$.
We also use $\Delta t_k=t_{k+1}-t_k$, $\mathbf l_k=\mathbf l(t_k)$, and $\mathbf l_{k+1}=\mathbf l(t_{k+1})$.
To establish notation for the subsequent derivation, we record the quantities associated with one PIE observation as
\begin{equation}
\mathcal P_k =
\left(
\mathbf u,
t_k,
t_{k+1},
p_{k+1},
\Delta t_k,
\mathbf l_k,
\mathbf l_{k+1}
\right).
\label{eq:pie}
\end{equation}
The resulting PIE record contains the event-pair information and corresponding light directions, without explicitly including the unknown contrast threshold $C$.

\subsection{Physical Irradiance Event Graph Neural Network}
\label{sec:network}
For each PIE $\mathcal P_k$, we define its PIE Feature (PIEF) as
\begin{equation}
\phi_k =
\frac{p_{k+1}}{\Delta t_k}.
\label{eq:pief}
\end{equation}
This scalar is the signed event rate.
It keeps event timing and polarity, and it does not require the unknown threshold scale $C$.

We treat each PIE as one PIE-GNN node.
Let $\tau_k=(t_k+t_{k+1})/2$ be its midpoint time and let $\bar{\mathbf u}_k$ be its normalized pixel coordinate.
The node coordinate is
\begin{equation}
\mathbf q_k =
\left[
\bar{\mathbf u}_k^{\top},
\gamma\tau_k
\right]^{\top},
\label{EQ:pie_node_coord}
\end{equation}
where $\gamma$ is a time scale.
We build the graph used by the PIE-GNN as
\begin{equation}
\mathcal G_{\mathrm{pie}}=(\mathcal V_{\mathrm{pie}},\mathcal E_{\mathrm{pie}}),
\qquad
\mathcal V_{\mathrm{pie}}=\{\mathcal P_k\}.
\label{EQ:pie_gnn_graph}
\end{equation}
The edge set $\mathcal E_{\mathrm{pie}}$ connects PIE nodes that are close in image space and time.
We add a directed edge $e_{ij}$ when
\begin{equation}
\left\|\bar{\mathbf u}_i-\bar{\mathbf u}_j\right\|_{\infty}
\le r_{\mathrm{pie}},
\qquad
0 < \tau_j-\tau_i \le \Delta\tau_{\max}.
\label{EQ:pie_edge_rule}
\end{equation}
Here $r_{\mathrm{pie}}$ is the spatial search radius.
$\Delta\tau_{\max}$ is the temporal window.
For each edge, we define the edge feature as
\begin{equation}
\mathbf a_{ij}
=
\frac{\mathbf q_j-\mathbf q_i}{2\rho_{\mathrm{pie}}}
+\frac{1}{2}\mathbf 1,
\qquad e_{ij}\in\mathcal E_{\mathrm{pie}} .
\label{EQ:pie_edge_feature}
\end{equation}
Here, $\rho_{\mathrm{pie}}$ is the radius used to normalize the relative coordinate, and $\mathbf 1\in\mathbb R^3$ denotes the all-ones vector.

For graph feature extraction, each PIE node uses the PIEF and the light-pair geometry as its base feature:
\begin{equation}
\mathbf x_k=
\left[
\phi_k,
\mathbf l_k^{\top},
\mathbf l_{k+1}^{\top}
\right]^{\top}.
\label{EQ:graph_node_feature}
\end{equation}
The pixel coordinate is used as an additional input for graph convolution and neighbor search.
A spline graph convolution extracts a contextual PIE feature from $\mathcal G_{\mathrm{pie}}$:
\begin{equation}
\mathbf h_k=
f_{\mathrm{obs}}
\left(
\mathrm{GConv}_{\mathrm{pie}}(\mathbf x_k,\bar{\mathbf u}_k,\mathcal E_{\mathrm{pie}},\{\mathbf a_{ij}\})
\right).
\label{EQ:pie_embedding}
\end{equation}
Here $\mathrm{GConv}_{\mathrm{pie}}$ denotes the PIE-GNN graph convolution, using $\mathcal E_{\mathrm{pie}}$ and its edge features. $f_{\mathrm{obs}}$ is an MLP that maps the graph-convolution output to a PIE embedding. Thus, $\mathbf h_k$ keeps the signed-rate measurement, the local light-pair cues, and context from nearby PIEs.

Let $\mathcal N(\mathbf u)=\{k\mid \mathcal P_k \text{ is observed at pixel } \mathbf u\}$.
This is the set of PIE nodes that belong to pixel $\mathbf u$.
The Reliability-Grading Attention (RGA) module predicts a reliability grade $w_k\in[0,1]$ for each PIE node.
It is described in Sec.~\ref{sec:LCA}.
We use these grades to summarize PIE embeddings at each pixel:
\begin{equation}
\bar{\mathbf h}_{\mathbf u}
=
\frac{\sum_{k\in\mathcal N(\mathbf u)}w_k\mathbf h_k}
{\sum_{k\in\mathcal N(\mathbf u)}w_k+\epsilon},
\qquad
\mathbf m_{\mathbf u}
=
\max_{k\in\mathcal N(\mathbf u)}
w_k\mathbf h_k .
\label{EQ:weighted_pool}
\end{equation}
Here $\epsilon$ avoids division by zero, and the max is taken element-wise.
The pixel feature is obtained by projecting the concatenated statistics:
\begin{equation}
\mathbf s_{\mathbf u}
=
f_{\mathrm{pool}}
\left(
[\bar{\mathbf h}_{\mathbf u},\mathbf m_{\mathbf u}]
\right).
\label{EQ:pixel_feature_after_lca}
\end{equation}

A lightweight pixel aggregator then refines these features over nearby valid pixels.
Let $\mathcal M(\mathbf u)$ denote the spatial neighbors of pixel $\mathbf u$.
The aggregator output is
\begin{equation}
\mathbf q_{\mathbf u}
=
F_{\mathrm{pix}}
\left(
\{\mathbf s_{\mathbf v}\mid \mathbf v\in\mathcal M(\mathbf u)\cup\{\mathbf u\}\}
\right).
\label{EQ:pixel_aggregator}
\end{equation}
More details of this pixel aggregation $F_{\mathrm{pix}}$ are provided in the appendix.
The normal head maps this refined pixel feature to the final normal:
\begin{equation}
\hat{\mathbf n}_{\mathbf u}
=
\mathrm{Norm}
\left(
f_{\mathrm{normal}}(\mathbf q_{\mathbf u})
\right),
\label{EQ:final_normal}
\end{equation}
where $f_{\mathrm{pool}}$ and $f_{\mathrm{normal}}$ are MLPs, and $\mathrm{Norm}(\cdot)$ denotes $L_2$ normalization.

\begin{figure}[t]
  \centering
  \includegraphics[width= 0.95\linewidth]{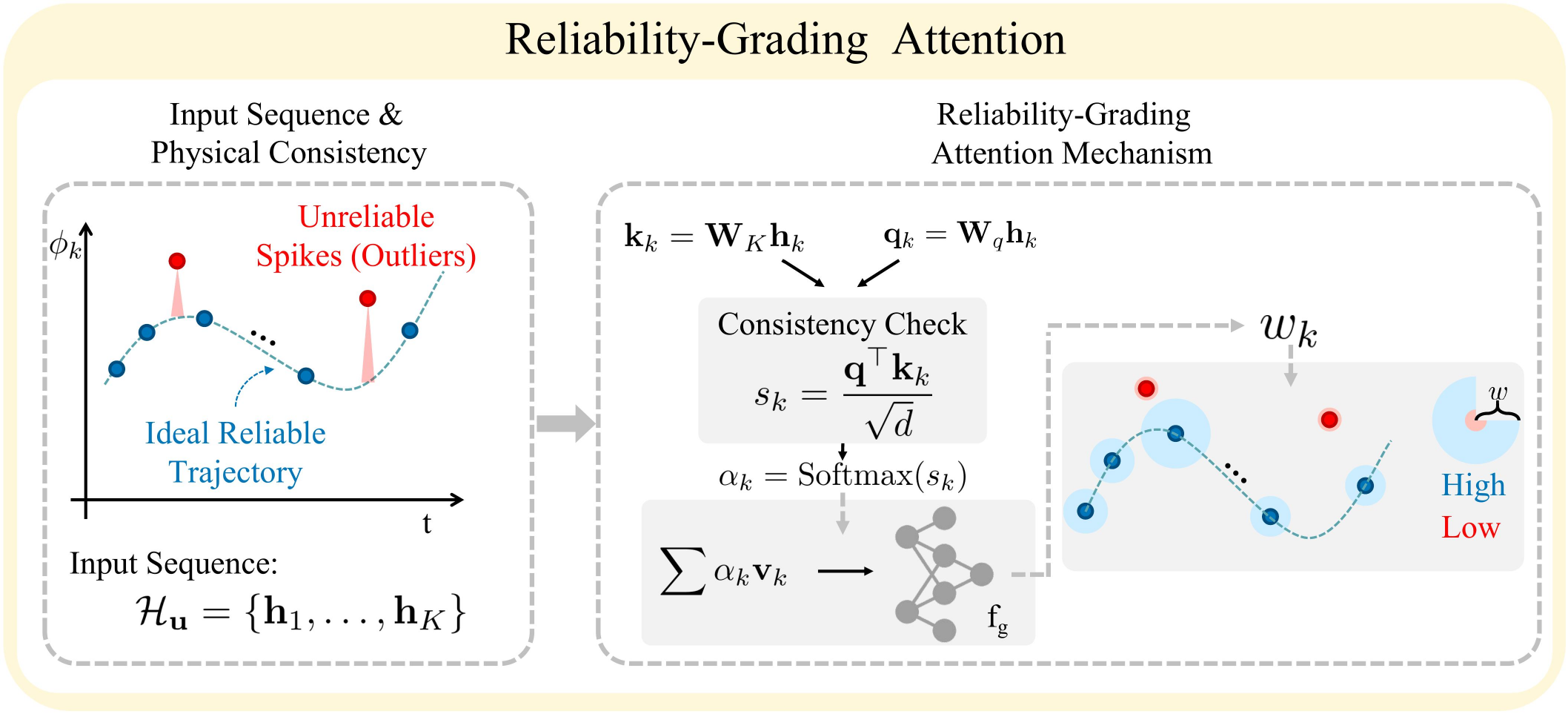}
  \caption{ The Reliability-Grading Attention module scores each event-pair observation. It detects spikes in the PIE sequence and predicts a confidence weight for each observation. The weighted observations are then aggregated into pixel features.}
  \label{LCA}
\end{figure}

\subsection{Reliability-Grading Attention}
\label{sec:LCA}
Eq.~\ref{calculate} shows that the PIEF is proportional to the measured log-image-irradiance change, up to the unknown scale $C$.
Under ideal Lambertian reflectance and smoothly varying illumination, PIEF observations at the same pixel are expected to vary consistently with their corresponding light pairs.
In practice, the reliability of PIE observations can vary with local appearance, illumination geometry, and sensor noise.
Less reliable observations may appear as abrupt deviations from this same-pixel temporal trend, as illustrated in Fig.~\ref{LCA}.
We propose RGA to assign each PIE a soft reliability grade for subsequent aggregation.

For a reliable PIE, the measured event contrast should agree with the Lambertian change predicted by the two light directions.
Substituting Eq.~\ref{EQ:log_lambertian} into Eq.~\ref{EQ:event_pair_trigger} gives
\begin{equation}
\ln\left(\mathbf n_{\mathbf u}^{\top}\mathbf l_{k+1}\right)
-
\ln\left(\mathbf n_{\mathbf u}^{\top}\mathbf l_k\right)
\approx p_{k+1}C .
\end{equation}
Exponentiating both sides gives
\begin{equation}
\frac{\mathbf n_{\mathbf u}^{\top}\mathbf l_{k+1}}
{\mathbf n_{\mathbf u}^{\top}\mathbf l_k}
\approx
\exp(p_{k+1}C).
\end{equation}
This step uses the discrete relation between two adjacent events rather than the continuous-time derivative.
This leads to the discrete pairwise constraint
\begin{equation}
\mathbf n_{\mathbf u}^{\top}
\left(
\mathbf l_{k+1}
-
\exp(p_{k+1}C)\mathbf l_k
\right)
\approx 0 .
\label{EQ:eventps_pair_constraint}
\end{equation}
When a PIE is reliable, the left-hand side of Eq.~\ref{EQ:eventps_pair_constraint} should be close to zero.
A large magnitude therefore indicates a potentially unreliable PIE observation.

In synthetic data, the ground-truth normal $\mathbf n_{\mathrm{gt},\mathbf u}$ and the threshold $C$ are known.
We use Eq.~\ref{EQ:eventps_pair_constraint} only to identify unreliable PIEs for analysis and comparison.
We define
\begin{equation}
\mathbf z_k
=
\mathbf l_{k+1}
-
\exp(p_{k+1}C)\mathbf l_k,
\qquad
r_k
=
\frac{
\left|\mathbf n_{\mathrm{gt},\mathbf u}^{\top}\mathbf z_k\right|
}{
\|\mathbf z_k\|_2+\epsilon
}.
\label{EQ:gt_unreliable_pie_residual}
\end{equation}
The score $r_k$ measures how much one PIE violates the discrete pairwise constraint.
A small $r_k$ means this PIE agrees with the Lambertian prediction.
A large $r_k$ means this PIE does not agree with the prediction, so it is treated as unreliable.
We also handle shadow and grazing lighting separately.
In these cases, $\mathbf n_{\mathrm{gt},\mathbf u}^{\top}\mathbf l_k$ or $\mathbf n_{\mathrm{gt},\mathbf u}^{\top}\mathbf l_{k+1}$ is non-positive or close to zero, so the Lambertian log-image-irradiance equation is invalid or unstable.
We therefore mark these PIEs as unreliable without using $r_k$.
The remaining PIEs are treated as GT-identified reliable PIEs.
These labels are used only for visualization, diagnostics, and comparison.
They are not model inputs at test time.

RGA suppresses unreliable observations by predicting a confidence weight for each PIE.
In our implementation, RGA uses a soft-gating attention.
It assigns an individual confidence weight to each PIE in the same-pixel sequence.
Fig.~\ref{LCA} shows this process on the time-ordered PIE sequence of one pixel.
Given a pixel $\mathbf u$, we
\noindent collect the time-ordered PIE embeddings
\begin{equation}
\mathcal H_{\mathbf u}
=
\{\mathbf h_k\mid k\in\mathcal N(\mathbf u)\}.
\label{EQ:pixel_embedding_sequence}
\end{equation}
The embeddings are sorted by time.

A small self-attention scorer reads the time-ordered embedding sequence and predicts an individual confidence weight for each PIE.
For each PIE embedding, we first form query, key, and value vectors:
\begin{equation}
\mathbf q_k=W_q\mathbf h_k,
\qquad
\mathbf k_k=W_k\mathbf h_k,
\qquad
\mathbf v_k=W_v\mathbf h_k .
\end{equation}
The attention score for PIE $k$ is computed from all PIEs at the same pixel:
\begin{equation}
\alpha_{kj}
=
\mathrm{softmax}_{j\in\mathcal N(\mathbf u)}
\left(
\frac{\mathbf q_k^{\top}\mathbf k_j}{\sqrt d}
\right),
\qquad
\mathbf y_k
=
\sum_{j\in\mathcal N(\mathbf u)}
\alpha_{kj}\mathbf v_j .
\label{EQ:lca_attention}
\end{equation}
The RGA confidence weight is then
\begin{equation}
w_k=\sigma(f_g(\mathbf y_k)).
\label{EQ:soft_gate}
\end{equation}
Here $f_g$ maps each attention output to a scalar logit, and $\sigma(\cdot)$ is the sigmoid function.
The sigmoid output lets multiple PIEs remain active at one pixel.
RGA reduces the contribution of PIEs assigned lower reliability grades.
The weights are used in Eq.~\ref{EQ:weighted_pool} to form the pixel feature.

\subsection{Training Strategy}
\label{sec:training}
We train the network with direct normal supervision.
Let $\Omega$ be the set of valid training pixels.
The main loss is the mean angular error on the final normal:
\begin{equation}
\mathcal L_{\mathrm{normal}}
=
\frac{1}{|\Omega|}
\sum_{\mathbf u\in\Omega}
\arccos
\left(
\hat{\mathbf n}_{\mathbf u}^{\top}\mathbf n_{\mathrm{gt},\mathbf u}
\right).
\label{EQ:normal_loss}
\end{equation}


At test time, the model only needs event pairs and light directions.

\section{Experiment}
\subsection{Implementation Details}

\paragraph{Method Implementation}
We implement PIE-PS in PyTorch.
The physics-only direct solver is implemented in NumPy.
For each pixel, it optimizes Eq.~\ref{EQ:direct_matching_objective} independently.
We use exact SVD for initialization and run five IRLS iterations with two tangent-plane Gauss-Newton steps.
Pixels with fewer than five PIE observations are not evaluated.
For PIE-GNN, we use a 7-D node feature, a 64-D PIE embedding, and a 128-D graph hidden feature. The neighborhood search modules use custom CUDA kernels.
For the PIE graph, we set the spatial search radius to $r_{\mathrm{pie}}=0.05$.
PIE times are normalized to $[0,1]$ within each sequence, and we use the full normalized range as the temporal window, so $\Delta\tau_{\max}=1$.
The temporal scale in Eq.~\ref{EQ:pie_node_coord} is $\gamma=0.3$.
For edge-feature normalization, we set $\rho_{\mathrm{pie}}=2\lfloor r_{\mathrm{pie}}W+2\rfloor/W$, where $W$ is the image width.
The pixel aggregator implementation is detailed in the supplementary material.
The RGA scorer is a two-layer Transformer with four attention heads and a 64-D hidden feature.
We optimize the network with Adam using the normal-estimation loss in Sec.~\ref{sec:training}.
Training takes about 3.4 hours for 200 epochs on a single NVIDIA H100 PCIe GPU.

\begin{figure}[t]
  \centering
  \includegraphics[width=0.95\linewidth]{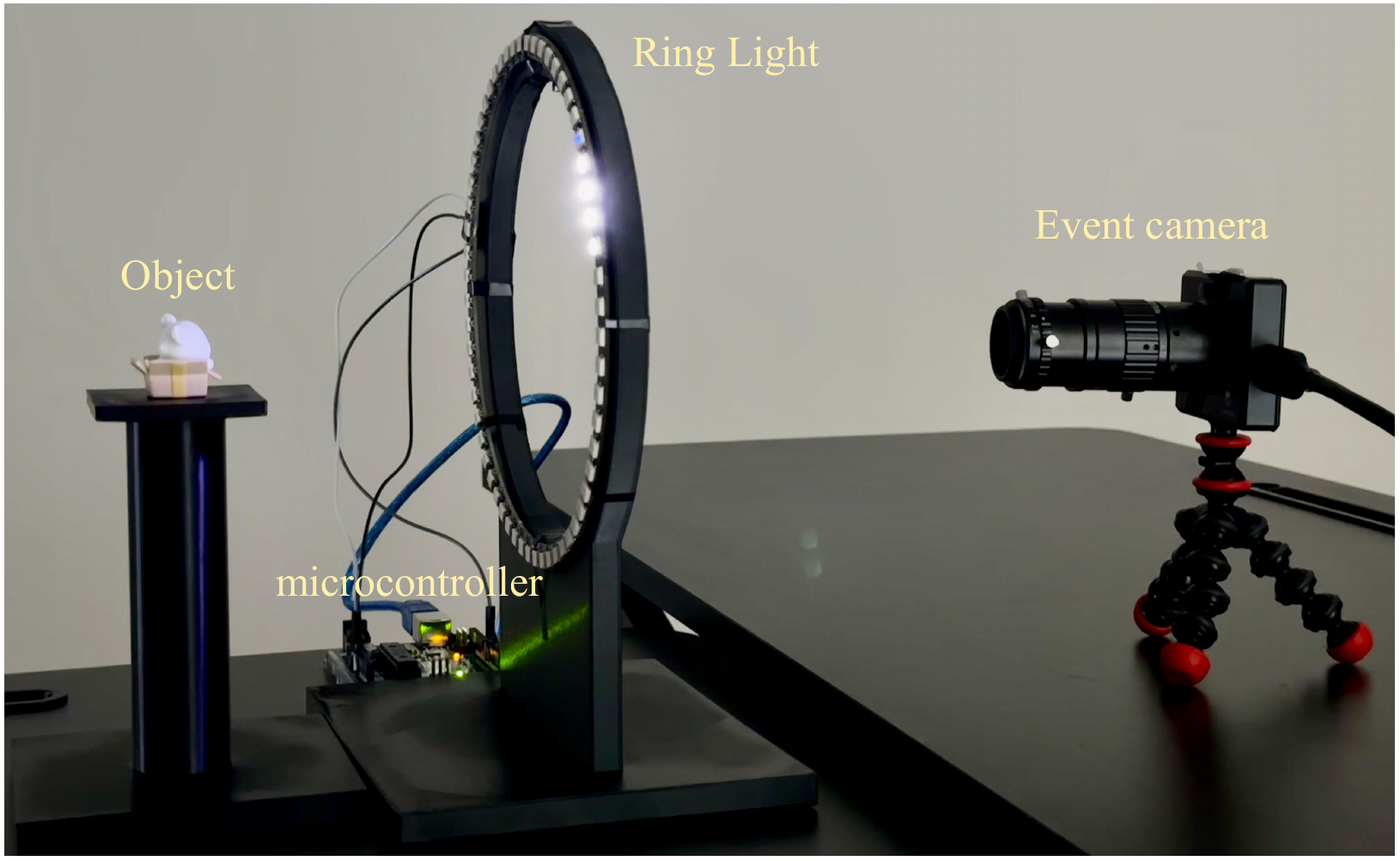}
  \caption{Prototype acquisition system used for real-world event-based photometric stereo.}
  \Description{A prototype capture setup with an event camera, a macro lens, and a programmable ring light for moving illumination.}
  \label{fig:prototype}
\end{figure}

\begin{figure*}[p]
  \centering
  \includegraphics[width=0.95\textwidth]{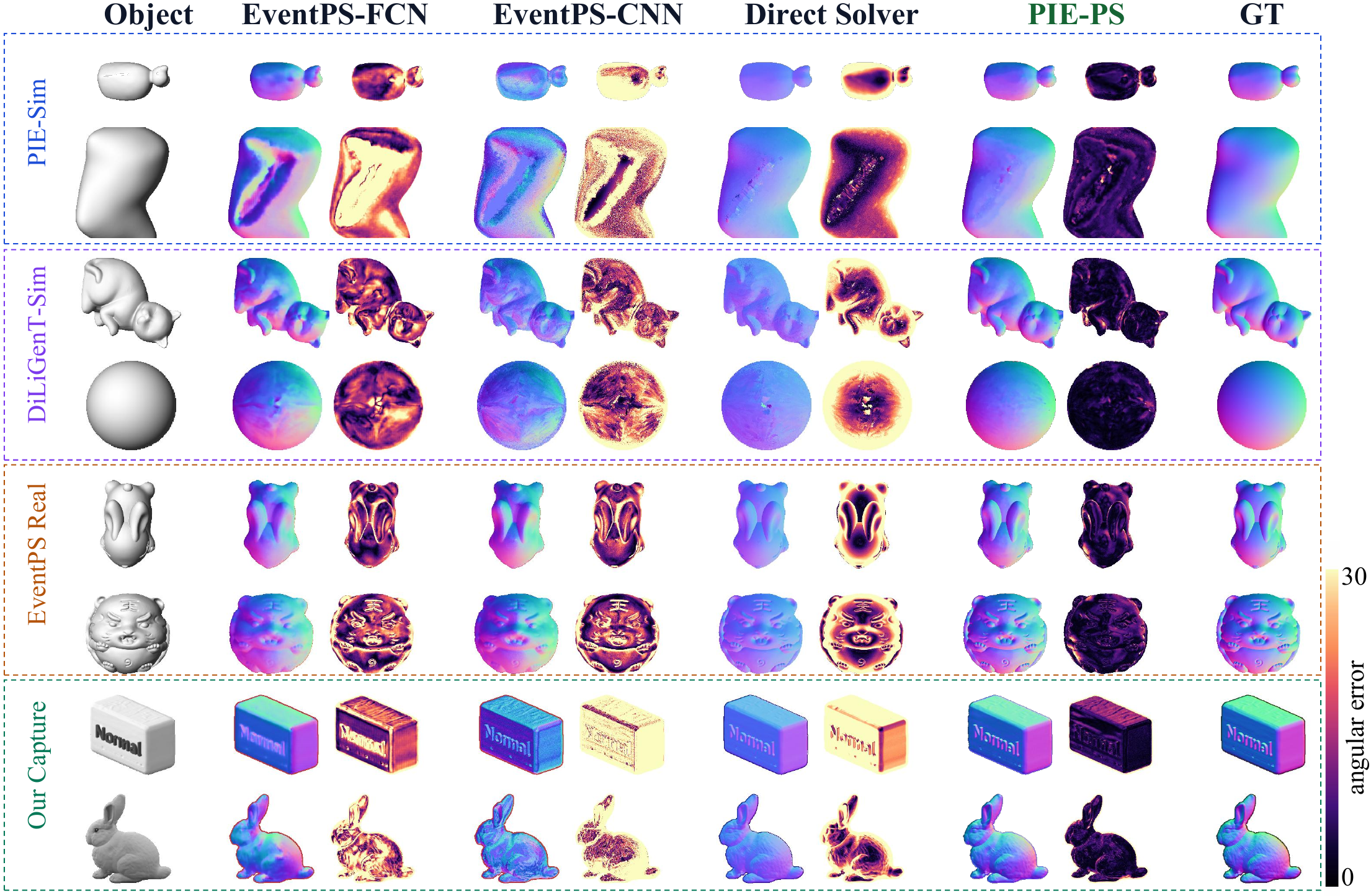}
  \caption{Qualitative comparison across synthetic and real scenes. Each row shows the object, the normal prediction and angular error map of each method, and the ground-truth normal. We include PIE-Sim, DiLiGenT-Sim, EventPS real scenes, and our captured scenes. Error maps use the same $0^\circ$--$30^\circ$ color range, where darker colors indicate lower error.}
  \Description{A qualitative comparison figure with eight rows. The rows cover PIE-Sim, DiLiGenT-Sim, EventPS real scenes, and captured scenes. Each row compares EventPS-FCN, EventPS-CNN, Direct Solver, and PIE-PS using normal maps and angular error maps.}
  \label{fig:qualitative_eval}
  \vspace{-6pt}
\end{figure*}

\begin{figure*}[p]
  \centering
  \includegraphics[width=0.95\textwidth]{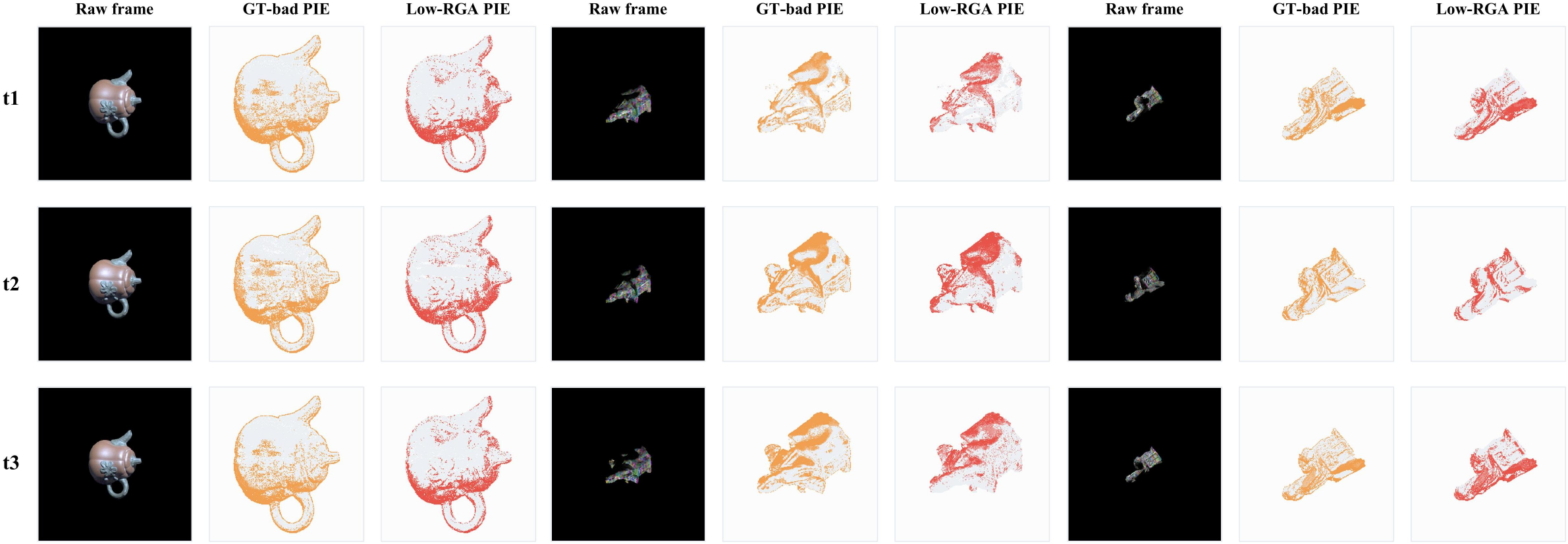}
  \caption{RGA reliability visualization. Each object is shown at three time slices. For each time slice, we show the raw frame, the GT-identified unreliable PIE ratio, and the low-RGA PIE ratio. GT-identified unreliable PIEs are computed with ground-truth normals and known simulation thresholds only for analysis. The figure label \emph{GT-bad} denotes these GT-identified unreliable PIEs. Low-RGA maps are produced by the learned RGA scorer.}
  \Description{A visualization with one DiLiGenT object and two PIE-Sim objects. Each object has three time rows and columns for raw frame, GT-identified unreliable PIE, and low-RGA PIE.}
  \label{fig:lca_reliability_vis}
  \vspace{-6pt}
\end{figure*}

\begin{table*}[!t]
\centering
\caption{Quantitative comparison on DiLiGenT-Sim. MAE is reported in degrees, and lower is better.}
\scriptsize
\resizebox{0.95\textwidth}{!}{
\begin{tabular}{lcccccccccc}
\toprule
\textbf{Method} & \textsc{Ball} & \textsc{Buddha} & \textsc{Cat} & \textsc{Cow} & \textsc{Goblet} & \textsc{Harvest} & \textsc{Pot1} & \textsc{Pot2} & \textsc{Reading} & \textbf{Average} \\
\midrule
EventUPS & 8.86 & 12.21 & 10.78 & 18.58 & 12.97 & 23.32 & 8.70 & 15.13 & 16.71 & 14.14 \\
EventPS-OP & 10.99 & 18.73 & 12.74 & 26.51 & 18.43 & 36.06 & 13.78 & 15.75 & 24.61 & 19.73 \\
EventPS-FCN & 7.49 & 18.13 & 11.42 & 20.61 & 18.07 & 26.05 & 12.83 & 16.59 & 15.16 & 16.26 \\
EventPS-CNN & 10.44 & 16.79 & 11.88 & 20.60 & 16.44 & 25.26 & 12.93 & 15.54 & 18.19 & 16.45 \\
\midrule
Direct Solver & 15.36 & 15.09 & 13.75 & 14.55 & 15.62 & 20.07 & 13.59 & 14.67 & 17.72 & 15.60 \\
PIE-PS & \textbf{3.93} & \textbf{8.67} & \textbf{5.15} & \textbf{12.90} & \textbf{7.81} & \textbf{12.91} & \textbf{6.33} & \textbf{8.34} & \textbf{12.25} & \textbf{8.70} \\
\bottomrule
\end{tabular}%
}
\label{tab:results}
\end{table*}

\paragraph{Prototype System} To validate our method on real-world data, we constructed the prototype acquisition system shown in Fig.~\ref{fig:prototype}.
The imaging setup uses a CeleX-V event camera with a spatial resolution of $1280 \times 800$.
The camera is equipped with a Zhonglian Kechuang TM5028MP12 macro lens with a 50mm focal length, 1.1-inch format, and C-mount interface.
For illumination, we use a programmable ring light with 60 WS2812 5050 RGB LED beads.
An Arduino Uno R3 sequentially drives the LEDs and records the activation timestamps and light directions.
The lighting timestamps are manually synchronized with the event camera clock.
We calibrate the setup so that the camera optical center, ring light center, and target object are aligned on the same horizontal axis.

\subsection{Datasets}

To train PIE-PS and ensure fair comparison with prior methods, we generate a synthetic dataset, named PIE-Sim.
PIE-Sim follows the same simulation pipeline as EventPS~\cite{eventps}.
It uses objects from the Blobby~\cite{Blobby} and Sculpture~\cite{Sculpture} collections, rendered with random poses and appearance parameters following the EventPS setup.
We simulate event streams using ESIM~\cite{ESIM} under three lighting trajectories, including circle, hypotrochoid, and interpolated DiLiGenT.
Each PIE-Sim scene is generated with an independently sampled contrast threshold and simulation-noise realization, producing different event densities and inter-event intervals.
Each sequence contains 600 rendered frames.
PIE-Sim contains 77 training scenes and 30 test scenes with no object overlap between the two splits.
To test generalization to standard photometric-stereo geometry, we also introduce DiLiGenT-Sim.
It is built by simulating event streams on DiLiGenT~\cite{DiLiGenT)}.
DiLiGenT-Sim contains 9 test scenes.

For real-world evaluation, we use two real data sources.
First, we include the three real EventPS scenes~\cite{eventps}.
Second, we capture three 3D-printed objects with our prototype system, including Bunny, Rectangle, and Flower.
We compute GT normals from the corresponding 3D CAD models and align them to the sensor observation using the event accumulation image~\cite{PS-EIP}.
The alignment is performed with MeshLab's mutual information registration filter~\cite{meshlab}, following the method in~\cite{DiLiGenT)}.

\subsection{Quantitative Evaluation}

Table~\ref{tab:results} reports the MAE on DiLiGenT-Sim.
For baselines, we use the published EventUPS and EventPS results on the same scene set.
EventUPS has no public code, and PS-EIP is omitted because its code and benchmark data are unavailable.
The direct solver is a physics-only baseline that optimizes Eq.~\ref{EQ:direct_matching_objective}.
PIE-PS achieves an average MAE of $8.70^\circ$.
It is $5.44^\circ$ lower than the best published baseline EventUPS and $6.90^\circ$ lower than the direct solver.

\begin{table}[H]
\centering
\caption{PIE-Sim results by object family and lighting trajectory. MAE is reported in degrees.}
\footnotesize
\resizebox{\linewidth}{!}{
\begin{tabular}{lccccccc}
\toprule
\multirow{2}{*}{\textbf{Method}} & \multicolumn{3}{c}{\textbf{Blobby}} & \multicolumn{3}{c}{\textbf{Sculpture}} & \multirow{2}{*}{\textbf{Avg.}} \\
\cmidrule(lr){2-4}\cmidrule(lr){5-7}
& Circle & Hypo. & DiLiGenT & Circle & Hypo. & DiLiGenT & \\
\midrule
EventPS-FCN & 28.83 & 16.83 & 21.73 & 31.63 & 26.43 & 29.59 & 25.84 \\
EventPS-CNN & 43.50 & 42.11 & 26.08 & 48.31 & 46.85 & 28.08 & 39.15 \\
Direct Solver & 27.08 & 17.24 & 13.02 & 26.60 & 24.68 & 22.67 & 21.88 \\
PIE-PS & \textbf{10.68} & \textbf{6.41} & \textbf{4.81} & \textbf{17.17} & \textbf{12.09} & \textbf{12.77} & \textbf{10.66} \\
\bottomrule
\end{tabular}%
}
\label{tab:piesim_results}

\end{table}

Table~\ref{tab:piesim_results} reports a breakdown by object family and lighting trajectory on PIE-Sim.
We evaluate EventPS-FCN and EventPS-CNN using their released models.
PIE-PS gives the lowest error in every group.
Its average MAE is $10.66^\circ$, which is $11.22^\circ$ lower than the direct solver and $15.18^\circ$ lower than EventPS-FCN.

\begin{table}[H]
\centering
\caption{Real-world results on EventPS scenes and our captures. MAE is reported in degrees.}
\footnotesize
\resizebox{\linewidth}{!}{
\begin{tabular}{lccccccc}
\toprule
\multirow{2}{*}{\textbf{Method}} & \multicolumn{3}{c}{\textbf{EventPS Real}} & \multicolumn{3}{c}{\textbf{Captured}} & \multirow{2}{*}{\textbf{Avg.}} \\
\cmidrule(lr){2-4}\cmidrule(lr){5-7}
& Real-1 & Real-2 & Real-3 & Bunny & Rectangle & Flower & \\
\midrule
EventPS-FCN & 15.79 & 19.65 & 9.42 & 26.14 & 18.70 & 35.76 & 20.91 \\
EventPS-CNN & 17.65 & 20.16 & 14.85 & 48.78 & 48.11 & 43.14 & 32.12 \\
Direct Solver & 21.08 & 21.10 & 20.50 & 23.46 & 32.22 & 25.02 & 23.90 \\
PIE-PS & \textbf{6.51} & \textbf{6.09} & \textbf{5.92} & \textbf{12.69} & \textbf{11.52} & \textbf{19.57} & \textbf{10.38} \\
\bottomrule
\end{tabular}%
}
\label{tab:real_results}
\end{table}

Table~\ref{tab:real_results} reports the MAE on six real-world scenes with aligned ground-truth normals.
We compare the released EventPS networks, the direct solver, and PIE-PS on the same scenes.
PIE-PS has the best result on all six scenes.
Its average MAE is $10.38^\circ$, which is $10.53^\circ$ lower than EventPS-FCN and $13.52^\circ$ lower than the direct solver.

We additionally evaluate PIE-PS under three CeleX SDK sensitivity settings, with detailed statistics reported in the supplementary material.

\subsection{Qualitative Evaluation}

Fig.~\ref{fig:qualitative_eval} compares reconstruction quality on both synthetic and real data.
EventPS-FCN and EventPS-CNN often recover the coarse object shape, but their error maps show large local errors, especially on high-curvature regions and real captures.
The direct solver gives a useful physics-only baseline, but it produces noisy normals because it solves each pixel independently and cannot reject unreliable event pairs.
In contrast, PIE-PS gives smoother and more accurate normals across the four settings.
The improvement is visible in the darker error maps and is consistent with the MAE values shown in the figure.

Fig.~\ref{fig:lca_reliability_vis} visualizes the RGA reliability weights.
The orange maps show GT-identified unreliable PIEs.
The red maps show PIEs assigned low RGA weights.
The low-RGA regions overlap more with GT-identified unreliable regions than with reliable regions.
This supports using RGA as a soft reliability-weighting module.

\section{Ablation Study}
We analyze PIEF and RGA on PIE-Sim.
For PIEF, we remove the signed event-rate input from each PIE node, while keeping the light-pair geometry and the rest of the network unchanged.
For RGA, we remove learned reliability weights and pool PIE observations directly.
All results are reported as MAE in degrees.

\begin{table}[H]
\centering
\caption{Ablation results on PIE-Sim. MAE is reported in degrees.}
\footnotesize
\setlength{\tabcolsep}{5pt}
\begin{tabular}{lccc}
\toprule
Metric & PIE-PS & w/o PIEF rate & w/o RGA \\
\midrule
Avg. MAE & \textbf{10.66} & 14.52 & 11.99 \\
\bottomrule
\end{tabular}
\label{tab:ablation_components}
\end{table}

\setcounter{figure}{6}
\begin{figure}[H]
  \centering
  \includegraphics[width=0.78\linewidth]{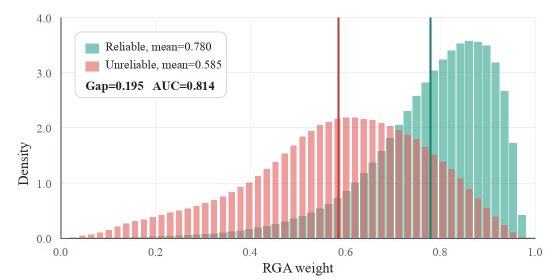}
  \caption{RGA weight distribution on PIE-Sim. GT-identified unreliable PIEs receive lower weights on average, supporting RGA as a soft reliability-weighting module.}
  \Description{A histogram comparing RGA weight distributions for reliable and unreliable PIEs on PIE-Sim. Reliable PIEs have a higher mean weight than unreliable PIEs.}
\label{fig:lca_weight_histogram}
\end{figure}

\balance
Table~\ref{tab:ablation_components} summarizes the contribution of the main PIE-PS components.
PIEF gives the main photometric signal, and removing it increases the average MAE from $10.66^\circ$ to $14.52^\circ$.
This shows that light-pair geometry alone is not enough, because it does not include the signed event response caused by the light change.
Removing RGA also increases the average MAE to $11.99^\circ$, confirming the value of learned reliability weighting.
Fig.~\ref{fig:lca_weight_histogram} further shows that RGA tends to assign lower weights to unreliable PIEs rather than weighting observations randomly.
This supports its role as a learned reliability weighting module.

\section{Conclusion and Discussion} 
In this paper, we presented PIE-PS, a physics-embedded framework that uses Physical Irradiance Events (PIEs) and a Physical Irradiance Event Graph Neural Network (PIE-GNN) to recover dense surface normals from event streams. Reliability-Grading Attention (RGA) assigns soft reliability grades to PIE observations before pixel aggregation.

Our method still assumes calibrated light directions. Strong specular highlights or complex inter-reflections can reduce the reliability of PIE observations and degrade normal estimation. In future work, we will extend the model to uncalibrated lighting and more general reflectance.

\begin{acks}
This work was supported in part by the National Natural Science Foundation of China (Grant Nos.~62572212 and 62506084), the Science and Technology Development Plan of Jilin Province (Grant No.~20260203049SF), and the Fundamental Research Funds for the Central Universities. This work was also supported by the High-Level Talents Innovation Team Project of the Guangdong-Macao In-Depth Cooperation Zone in Hengqin (Grant No.~2630004018939).
\end{acks}

\clearpage



\bibliographystyle{ACM-Reference-Format}
\bibliography{sample-base}

@String{Computer = "{IEEE} Computer" }

@String{Springer = "Springer-Verlag" }

@article{woodham1980photometric,
  title={Photometric method for determining surface orientation from multiple images},
  author={Woodham, Robert J},
  journal={Optical engineering},
  volume={19},
  number={1},
  pages={139--144},
  year={1980},
  publisher={SPIE}
}

@inproceedings{op1,
  title={Robust photometric stereo using sparse regression},
  author={Ikehata, Satoshi and Wipf, David and Matsushita, Yasuyuki and Aizawa, Kiyoharu},
  booktitle={2012 IEEE Conference on Computer Vision and Pattern Recognition},
  pages={318--325},
  year={2012},
  organization={IEEE}
}

@inproceedings{op2,
  title={Robust photometric stereo via low-rank matrix completion and recovery},
  author={Wu, Lun and Ganesh, Arvind and Shi, Boxin and Matsushita, Yasuyuki and Wang, Yongtian and Ma, Yi},
  booktitle={Asian conference on computer vision},
  pages={703--717},
  year={2010},
  organization={Springer}
}

@inproceedings{non-lam1,
  title={Shadowcuts: Photometric stereo with shadows},
  author={Chandraker, Manmohan and Agarwal, Sameer and Kriegman, David},
  booktitle={2007 IEEE Conference on Computer Vision and Pattern Recognition},
  pages={1--8},
  year={2007},
  organization={IEEE}
}

@article{non-lam2,
  title={Median photometric stereo as applied to the segonko tumulus and museum objects},
  author={Miyazaki, Daisuke and Hara, Kenji and Ikeuchi, Katsushi},
  journal={International Journal of Computer Vision},
  volume={86},
  number={2},
  pages={229--242},
  year={2010},
  publisher={Springer}
}

@article{deep1,
  title={Summary study of data-driven photometric stereo methods},
  author={Zheng, Qian and Shi, Boxin and Pan, Gang},
  journal={Virtual Reality \& Intelligent Hardware},
  volume={2},
  number={3},
  pages={213--221},
  year={2020},
  publisher={Elsevier}
}

@inproceedings{deep2,
  title={Self-calibrating deep photometric stereo networks},
  author={Chen, Guanying and Han, Kai and Shi, Boxin and Matsushita, Yasuyuki and Wong, Kwan-Yee K},
  booktitle={Proceedings of the IEEE/CVF conference on computer vision and pattern recognition},
  pages={8739--8747},
  year={2019}
}

@inproceedings{deep3,
  title={Deep photometric stereo network},
  author={Santo, Hiroaki and Samejima, Masaki and Sugano, Yusuke and Shi, Boxin and Matsushita, Yasuyuki},
  booktitle={Proceedings of the IEEE international conference on computer vision workshops},
  pages={501--509},
  year={2017}
}

@inproceedings{deep4,
  title={SPLINE-Net: Sparse photometric stereo through lighting interpolation and normal estimation networks},
  author={Zheng, Qian and Jia, Yiming and Shi, Boxin and Jiang, Xudong and Duan, Ling-Yu and Kot, Alex C},
  booktitle={Proceedings of the IEEE/CVF International Conference on Computer Vision},
  pages={8549--8558},
  year={2019}
}

@inproceedings{deep5,
  title={Px-net: Simple and efficient pixel-wise training of photometric stereo networks},
  author={Logothetis, Fotios and Budvytis, Ignas and Mecca, Roberto and Cipolla, Roberto},
  booktitle={Proceedings of the IEEE/CVF international conference on computer vision},
  pages={12757--12766},
  year={2021}
}

@inproceedings{SDPS,
  title={Self-calibrating photometric stereo by neural inverse rendering},
  author={Li, Junxuan and Li, Hongdong},
  booktitle={European Conference on Computer Vision},
  pages={166--183},
  year={2022},
  organization={Springer}
}

@article{DPSN,
  title={Deep photometric stereo networks for determining surface normal and reflectances},
  author={Santo, Hiroaki and Samejima, Masaki and Sugano, Yusuke and Shi, Boxin and Matsushita, Yasuyuki},
  journal={IEEE Transactions on Pattern Analysis and Machine Intelligence},
  volume={44},
  number={1},
  pages={114--128},
  year={2020},
  publisher={IEEE}
}

@inproceedings{li2024spin,
  title={Spin-up: Spin light for natural light uncalibrated photometric stereo},
  author={Li, Zongrui and Lu, Zhan and Yan, Haojie and Shi, Boxin and Pan, Gang and Zheng, Qian and Jiang, Xudong},
  booktitle={Proceedings of the IEEE/CVF Conference on Computer Vision and Pattern Recognition},
  pages={11905--11914},
  year={2024}
}

@article{event-based-vision,
  title={Event-based vision: A survey},
  author={Gallego, Guillermo and Delbr{\"u}ck, Tobi and Orchard, Garrick and Bartolozzi, Chiara and Taba, Brian and Censi, Andrea and Leutenegger, Stefan and Davison, Andrew J and Conradt, J{\"o}rg and Daniilidis, Kostas and others},
  journal={IEEE transactions on pattern analysis and machine intelligence},
  volume={44},
  number={1},
  pages={154--180},
  year={2020},
  publisher={IEEE}
}

@article{emvs,
  title={EMVS: Event-based multi-view stereo—3D reconstruction with an event camera in real-time},
  author={Rebecq, Henri and Gallego, Guillermo and Mueggler, Elias and Scaramuzza, Davide},
  journal={International Journal of Computer Vision},
  volume={126},
  number={12},
  pages={1394--1414},
  year={2018},
  publisher={Springer}
}

@inproceedings{Esl,
  title={Esl: Event-based structured light},
  author={Muglikar, Manasi and Gallego, Guillermo and Scaramuzza, Davide},
  booktitle={2021 International Conference on 3D Vision (3DV)},
  pages={1165--1174},
  year={2021},
  organization={IEEE}
}

@inproceedings{wang2022evac3d,
  title={Evac3d: From event-based apparent contours to 3d models via continuous visual hulls},
  author={Wang, Ziyun and Chaney, Kenneth and Daniilidis, Kostas},
  booktitle={European conference on computer vision},
  pages={284--299},
  year={2022},
  organization={Springer}
}

@inproceedings{rudnev2023eventnerf,
  title={Eventnerf: Neural radiance fields from a single colour event camera},
  author={Rudnev, Viktor and Elgharib, Mohamed and Theobalt, Christian and Golyanik, Vladislav},
  booktitle={Proceedings of the IEEE/CVF Conference on Computer Vision and Pattern Recognition},
  pages={4992--5002},
  year={2023}
}

@article{han2024eventgs,
  title={Event-3dgs: Event-based 3d reconstruction using 3d gaussian splatting},
  author={Han, Hanqian and Li, Jianing and Wei, Henglu and Ji, Xiangyang},
  journal={Advances in Neural Information Processing Systems},
  volume={37},
  pages={128139--128159},
  year={2024}
}

@article{niu2025esvo2,
  title={Esvo2: Direct visual-inertial odometry with stereo event cameras},
  author={Niu, Junkai and Zhong, Sheng and Lu, Xiuyuan and Shen, Shaojie and Gallego, Guillermo and Zhou, Yi},
  journal={IEEE Transactions on Robotics},
  year={2025},
  publisher={IEEE}
}

@inproceedings{nerf1,
  title={Robust e-nerf: Nerf from sparse \& noisy events under non-uniform motion},
  author={Low, Weng Fei and Lee, Gim Hee},
  booktitle={Proceedings of the IEEE/CVF International Conference on Computer Vision},
  pages={18335--18346},
  year={2023}
}

@inproceedings{nerf2,
  title={Deformable neural radiance fields using rgb and event cameras},
  author={Ma, Qi and Paudel, Danda Pani and Chhatkuli, Ajad and Van Gool, Luc},
  booktitle={Proceedings of the IEEE/CVF International Conference on Computer Vision},
  pages={3590--3600},
  year={2023}
}

@article{EFPS-Net,
    doi = {10.1016/j.neunet.2023.08.009},
    url = {https://doi.org/10.1016%2Fj.neunet.2023.08.009},
    year = 2023,
    month = {aug},
    publisher = {Elsevier {BV}},
    author = {Wonjeong Ryoo and Giljoo Nam and Jae-Sang Hyun and Sangpil Kim},
    title = {Event fusion photometric stereo network},              
    journal = {Neural Networks}
}

@inproceedings{eventups,
  title={EventUPS: Uncalibrated Photometric Stereo Using an Event Camera},
  author={Liang, Jinxiu and Yu, Bohan and Yang, Siqi and Zhuang, Haotian and Ren, Jieji and Duan, Peiqi and Shi, Boxin},
  booktitle={Proceedings of the IEEE/CVF International Conference on Computer Vision},
  pages={7516--7525},
  year={2025}
}

@inproceedings{PS-EIP,
  title={PS-EIP: Robust Photometric Stereo Based on Event Interval Profile},
  author={Kitazawa, Kazuma and Aoto, Takahito and Ikehata, Satoshi and Takatani, Tsuyoshi},
  booktitle={Proceedings of the Computer Vision and Pattern Recognition Conference},
  pages={6241--6251},
  year={2025}
}

@inproceedings{eventps,
  title={Eventps: Real-time photometric stereo using an event camera},
  author={Yu, Bohan and Ren, Jieji and Han, Jin and Wang, Feishi and Liang, Jinxiu and Shi, Boxin},
  booktitle={Proceedings of the IEEE/CVF Conference on Computer Vision and Pattern Recognition},
  pages={9602--9611},
  year={2024}
}

@inproceedings{qi2018geonet,
  title={Geonet: Geometric neural network for joint depth and surface normal estimation},
  author={Qi, Xiaojuan and Liao, Renjie and Liu, Zhengzhe and Urtasun, Raquel and Jia, Jiaya},
  booktitle={Proceedings of the IEEE Conference on Computer Vision and Pattern Recognition},
  pages={283--291},
  year={2018}
}

@inproceedings{bae2024rethinking,
  title={Rethinking inductive biases for surface normal estimation},
  author={Bae, Gwangbin and Davison, Andrew J},
  booktitle={Proceedings of the IEEE/CVF Conference on Computer Vision and Pattern Recognition},
  pages={9535--9545},
  year={2024}
}

@inproceedings{rebecq2019events,
  title={Events-to-video: Bringing modern computer vision to event cameras},
  author={Rebecq, Henri and Ranftl, Ren{\'e} and Koltun, Vladlen and Scaramuzza, Davide},
  booktitle={Proceedings of the IEEE/CVF Conference on Computer Vision and Pattern Recognition},
  pages={3857--3866},
  year={2019}
}

@inproceedings{Blobby,
  title={Shape estimation in natural illumination},
  author={Johnson, Micah K and Adelson, Edward H},
  booktitle={CVPR 2011},
  pages={2553--2560},
  year={2011},
  organization={IEEE}
}

@article{Sculpture,
  title={Silnet: Single-and multi-view reconstruction by learning from silhouettes},
  author={Wiles, Olivia and Zisserman, Andrew},
  journal={arXiv preprint arXiv:1711.07888},
  year={2017}
}

@inproceedings{ESIM,
  title={Esim: an open event camera simulator},
  author={Rebecq, Henri and Gehrig, Daniel and Scaramuzza, Davide},
  booktitle={Conference on robot learning},
  pages={969--982},
  year={2018},
  organization={PMLR}
}

@inproceedings{meshlab,
  title={Meshlab: an open-source mesh processing tool.},
  author={Cignoni, Paolo and Callieri, Marco and Corsini, Massimiliano and Dellepiane, Matteo and Ganovelli, Fabio and Ranzuglia, Guido and others},
  booktitle={Eurographics Italian chapter conference},
  volume={2008},
  pages={129--136},
  year={2008},
  organization={Salerno}
}

@inproceedings{ge2025prm,
  title={Prm: Photometric stereo based large reconstruction model},
  author={Ge, Wenhang and Lin, Jiantao and Shen, Guibao and Feng, Jiawei and Hu, Tao and Xu, Xinli and Chen, Ying-Cong},
  booktitle={Proceedings of the IEEE/CVF International Conference on Computer Vision},
  pages={25009--25018},
  year={2025}
}

@inproceedings{lichy2022fast,
  title={Fast light-weight near-field photometric stereo},
  author={Lichy, Daniel and Sengupta, Soumyadip and Jacobs, David W},
  booktitle={Proceedings of the IEEE/CVF Conference on Computer Vision and Pattern Recognition},
  pages={12612--12621},
  year={2022}
}

@inproceedings{feng2025ae,
  title={AE-NeRF: Augmenting Event-Based Neural Radiance Fields for Non-ideal Conditions and Larger Scenes},
  author={Feng, Chaoran and Yu, Wangbo and Cheng, Xinhua and Tang, Zhenyu and Zhang, Junwu and Yuan, Li and Tian, Yonghong},
  booktitle={Proceedings of the AAAI Conference on Artificial Intelligence},
  volume={39},
  number={3},
  pages={2924--2932},
  year={2025}
}

@inproceedings{feng20254dgs,
  title={E-4dgs: High-fidelity dynamic reconstruction from the multi-view event cameras},
  author={Feng, Chaoran and Tang, Zhenyu and Yu, Wangbo and Pang, Yatian and Zhao, Yian and Zhao, Jianbin and Yuan, Li and Tian, Yonghong},
  booktitle={Proceedings of the 33rd ACM International Conference on Multimedia},
  pages={7356--7365},
  year={2025}
}

@article{baek2018polarimetricSVBRDF,
  title={Simultaneous Acquisition of Polarimetric {SVBRDF} and Normals},
  author={Baek, Seung-Hwan and Jeon, Daniel S. and Tong, Xin and Kim, Min H.},
  journal={ACM Transactions on Graphics},
  volume={37},
  number={6},
  articleno={268},
  pages={1--15},
  year={2018},
  doi={10.1145/3272127.3275018}
}

@article{hwang2022sparseEllipsometry,
  title={Sparse Ellipsometry: Portable Acquisition of Polarimetric {SVBRDF} and Shape with Unstructured Flash Photography},
  author={Hwang, Inseung and Jeon, Daniel S. and Mu{\~n}oz, Adolfo and Gutierrez, Diego and Tong, Xin and Kim, Min H.},
  journal={ACM Transactions on Graphics},
  volume={41},
  number={4},
  articleno={145},
  pages={1--14},
  year={2022},
  doi={10.1145/3528223.3530075}
}

@article{xu2025event3dSurvey,
  title={A Survey of 3D Reconstruction with Event Cameras},
  author={Xu, Chuanzhi and Zhou, Haoxian and Chen, Langyi and Chen, Haodong and Zhou, Ying and Chung, Vera Yuk Ying and Qu, Qiang and Cai, Weidong},
  journal={arXiv preprint arXiv:2505.08438},
  year={2025}
}










\end{document}